\documentclass[letterpaper]{article} 
\usepackage[submission]{aaai23}  
\usepackage{times}  
\usepackage{helvet}  
\usepackage{courier}  
\usepackage[hyphens]{url}  
\usepackage{graphicx} 
\usepackage{natbib}  
\usepackage{caption} 
\usepackage{algorithm}
\usepackage{algorithmic}
\usepackage{amsmath}
\usepackage{amssymb}
\usepackage{booktabs}
\usepackage{multirow}
\usepackage{pifont}
\usepackage{subfigure}
\usepackage{url}
\usepackage[colorlinks,linkcolor=red]{hyperref}
\usepackage{newfloat}
\usepackage{listings}
\DeclareCaptionStyle{ruled}{labelfont=normalfont,labelsep=colon,strut=off} 
\floatstyle{ruled}
\newfloat{listing}{tb}{lst}{}
\floatname{listing}{Listing}
\title{Do we really need temporal convolutions in action segmentation?}
\author{
    Dazhao Du\textsuperscript{\rm{1,2}},
    Bing Su\textsuperscript{\rm3}\thanks{\ \ Corresponding author},
    Yu Li\textsuperscript{\rm4},
    Zhongang Qi\textsuperscript{\rm5},
    Lingyu Si\textsuperscript{\rm{1,2}},
    Ying Shan\textsuperscript{\rm{5}}
}
\affiliations{
    \textsuperscript{\rm 1}Institute of Software Chinese Academy of Sciences \\ 
    \textsuperscript{\rm 2}University of Chinese Academy of Sciences\\ 
    \textsuperscript{\rm 3}Renmin University of China \\
    \textsuperscript{\rm 4}International Digital Economy Academy \\
    \textsuperscript{\rm 5}ARC Lab, Tencent PCG \\
    dudazhao20@mails.ucas.ac.cn, bingsu@ruc.edu.cn \\ liyu@idea.edu.cn, zhongangqi@tencent.com, lingyu@iscas.ac.cn, yingsshan@tencent.com
}

\usepackage{bibentry}

\begin{document}

\maketitle

\begin{abstract}
Action classification has made great progress, but segmenting and recognizing actions from long untrimmed videos remains a challenging problem. Most state-of-the-art methods focus on designing temporal convolution-based models, but the inflexibility of temporal convolutions and the difficulties in modeling long-term temporal dependencies restrict the potential of these models. Transformer-based models with adaptable and sequence modeling capabilities have recently been used in various tasks. However, the lack of inductive bias and the inefficiency of handling long video sequences limit the application of Transformer in action segmentation. In this paper, we design a pure Transformer-based model without temporal convolutions by incorporating temporal sampling, called Temporal U-Transformer (TUT). The U-Transformer architecture reduces complexity while introducing an inductive bias that adjacent frames are more likely to belong to the same class, but the introduction of coarse resolutions results in the misclassification of boundaries. We observe that the similarity distribution between a boundary frame and its neighboring frames depends on whether the boundary frame is the start or end of an action segment. Therefore, we further propose a boundary-aware loss based on the distribution of similarity scores between frames from attention modules to enhance the ability to recognize boundaries. Extensive experiments show the effectiveness of our model.
\end{abstract}

\section{Introduction}
\label{sec:intro}

\begin{figure}[t]
\centering
\includegraphics[width=0.4\textwidth]{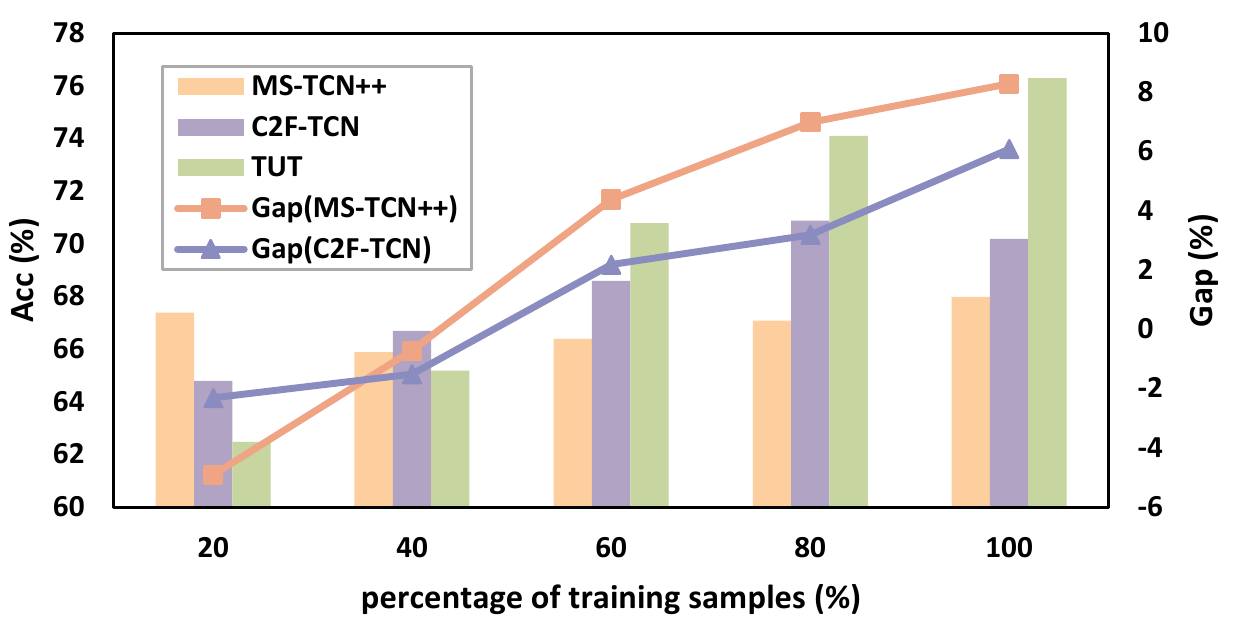} 
\caption{The accuracy metrics of our proposed TUT and two advanced TCN-based models, i.e., MS-TCN++ and C2F-TCN, are compared under different proportions of samples sampled from the original training set in the Breakfast dataset for training. The Gap curves refer to the performance gaps between TUT and TCN-based models.}
\label{fig:trainratio}
\end{figure}

Action recognition from videos is one of the most active tasks for video content understanding, which can be classified into two categories: classifying trimmed videos with a single activity and segmenting activities in untrimmed videos. The latter is also known as action segmentation. Although approaches based on various architectures have been proposed to improve the accuracy of video classification greatly, their performance is limited by the action segmentation task for untrimmed videos. 

Action segmentation can be treated as a frame-wise classification problem. Most of the previous deep learning methods adapt temporal convolutional networks (TCNs) as their backbones~\cite{lea2017temporal, farha2019ms, li2020ms}, which utilize temporal convolutions to capture the temporal relationships between different frames. However, TCNs need very deep layers to capture long-term dependencies and the optimal receptive field is hard to determine. The most popular TCN-based model, MS-TCN~\cite{farha2019ms}, adopts the strategy of doubling the dilation factor in 1D dilated convolution than the previous layer so that the receptive field grows exponentially with the number of layers, but Global2Local~\cite{gao2021global2local} has proven that there exist more effective receptive field combinations than this hand-designed pattern. Even different data distributions will result in different optimal receptive field combinations. Therefore, we need more flexible models that extract the dependency between frames from the data itself, instead of dilated convolutional structures with fixed weights and hand-designed patterns.

Transformer~\cite{Transformer} outperforms other deep models, e.g., RNNs~\cite{rnn} and TCNs~\cite{TCN}, in various fields due to its flexible modeling capabilities~\cite{kenton2019bert,girdhar2019video,dosovitskiy2020vit}. However, to our knowledge, there are few works utilizing Transformer to tackle the action segmentation task except ASFormer~\cite{ASFormer}. There exist two issues when applying Transformer to action segmentation. On the one hand, Transformer makes fewer assumptions about the structural bias of input data and thus requires larger amounts of data for training. However, limited by the difficulty of frame-wise annotations, most well-annotated datasets in the action segmentation task such as Breakfast~\cite{kuehne2014language} have only thousands of video samples, which are much smaller than the data scale in other fields~\cite{deng2009imagenet,kay2017kinetics}. On the other hand, the time and space complexities increase quadratically with the length of inputs. The untrimmed video samples consisting of thousands of frames are too long to be directly processed by the self-attention layer in Transformer. ASFormer~\cite{ASFormer} combines a sparse attention mechanism and temporal convolutions to tackle these two issues, but it is more like incorporating additional attention modules into MS-TCN. Therefore, it is still an open problem whether a pure Transformer-based model without temporal convolutions is suitable for action segmentation and how to make it work.

To be able to handle long videos, we first replace full attentions in vanilla Transformer with local attentions~\cite{beltagy2020longformer}, where each frame only attends to frames within the same local window. But local attention will reduce the receptive field so that the model still can not capture long-term dependencies. To this end, we introduce temporal sampling in each layer of the local-attended Transformer, resulting in a pure Transformer-based model without temporal convolutions. We call it \textit{\textbf{T}emporal \textbf{U}-\textbf{T}ransformer (TUT)} because temporal downsampling in the encoder and upsampling in the decoder are exploited to construct the temporal U-Net-like architecture~\cite{ron2015unet}. Temporal sampling not only increases the receptive field exponentially with the number of layers but also further reduces the complexity. Moreover, we find that the U-Transformer architecture is well suited for dense prediction tasks because it introduces multi-scale information and priors that adjacent frames are likely to belong to the same class, which compensates for the lack of sufficient training data on action segmentation. This has been demonstrated on another dense prediction task, i.e., semantic segmentation~\cite{xie2021segformer,cao2021swinunet,ji2021multi}. C2F-TCN~\cite{singhania2021coarse} also leverages the U-Net-like architecture for action segmentation, but it still uses temporal convolutions to build the model. 


However, because coarse-grained features are supplied into the decoder, the U-Transformer architecture exacerbates the misclassification of boundaries. When decoded, the frames close to the boundaries that are improperly encoded as coarse-grained features will be misclassified. To better perceive boundary information, we categorize the boundary frame in the video into two types: the start and the end frame, which represent the start and end of an action segment respectively, regardless of what the action is. Intuitively, the start frame should be more similar to the neighboring frames after it, while the end frame should be more similar to those before it, which corresponds to two different similarity distributions. We define the similarity distribution of a frame with its neighbors as the \textit{local-attention distribution} of the frame, which can be obtained from the local-attention module. We thereby introduce a boundary-aware loss by minimizing the distance between the local-attention distribution of the boundary frame with pre-defined prior distributions, which serves as a regularization to enforce the model to pay more attention to boundaries.

In summary, the contributions of the paper include: 
\begin{itemize}
\item For the first time, we propose a pure Transformer-based model without temporal convolutions for action segmentation. TUT incorporates local attention and temporal sampling into the Transformer, which has reduced complexity compared to the vanilla Transformer and breaks away from the constraints of temporal convolutions.
\item The superiority of our pure Transformer-based model TUT is shown in Fig.~\ref{fig:trainratio}. The performance of TUT gradually increases as the number of training samples increases. In contrast, the performance of two advanced TCN-Based models reaches saturation. Moreover, the gaps between TUT and them keep increasing.
\item Based on the distribution of inter-frame similarity scores from the attention module and boundary labels, we propose a distribution-based boundary-aware loss to enable our model to classify boundaries more accurately.
\end{itemize}

\begin{figure*}[t]
\centering
\includegraphics[width=0.8\textwidth]{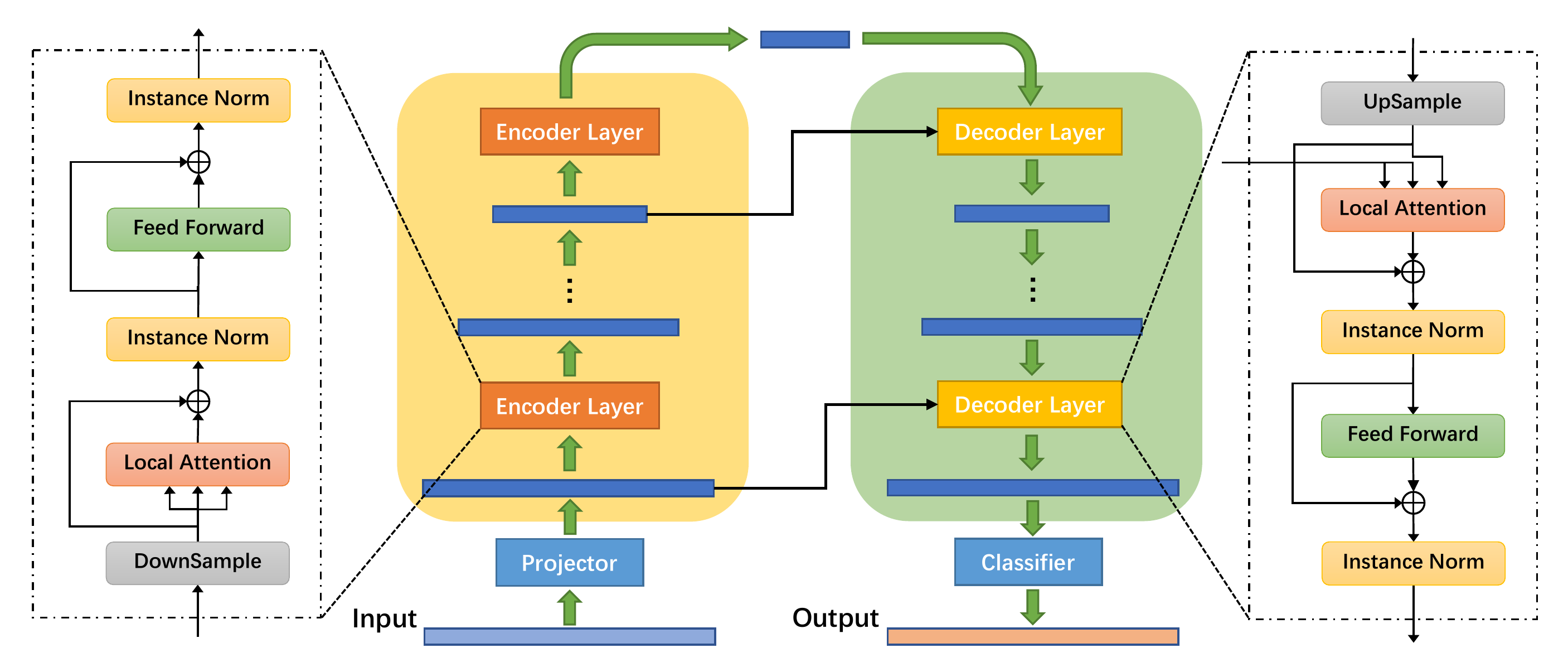} 
\caption{Overview of one stage in our TUT model. Since all stages have the same architecture, we only show one stage, which contains two main parts. The encoder stacks several encoder layers to capture temporal dependencies to generate high-level features. These features are passed into the decoder, where decoder layers are utilized to generate frame-wise features.}
\label{fig:model}
\end{figure*}

\section{Related work}
\label{sec:relatedwork}

{\bf Action segmentation} \quad 
Earlier approaches~\cite{rohrbach2012database,karaman2014fast} apply a sliding window and use non-maximum suppression to obtain the action segments. Some approaches employ linear dynamical systems~\cite{bhattacharya2014recognition} or Bayesian non-parametric models~\cite{cheng2014temporal} to model the changes in actions. Such models are difficult to capture the long-term dependencies within a large-scale context. To model the long-term temporal dynamics, graphical models and sequential models such as hidden Markov model (HMM)~\cite{kuehne2016end,kuehne2017weakly}, RNN~\cite{richard2017weakly,singh2016multi}, stochastic context-free grammar~\cite{vo2014stochastic}, and spatiotemporal CNN~\cite{lea2016segmental} are also used to perform action classification at each frame. Motivated by the success of WaveNet~\cite{van2016wavenet} in speech synthesis, many recent works are devoted to exploring multi-scale information with temporal convolution networks (TCNs) for action segmentation~\cite{lea2017temporal}. In TDRN~\cite{lei2018temporal}, deformable convolution is used to replace conventional convolution and residual connections are added to improve TCN. MS-TCN~\cite{farha2019ms} designs a multi-stage architecture which is a stack of multiple TCNs and proposes the truncated MSE loss to penalize over-segmentations. It gradually alleviates the over-segmentation issue in each refinement stage. Some researchers add additional modules to the MS-TCN, e.g., the boundary prediction module~\cite{BCN}, the bilinear pooling operation~\cite{zhang2019low} and the graph-based temporal reasoning module~\cite{huang2020improving}. MS-TCN++~\cite{li2020ms} replaces dilated temporal convolution layers in MS-TCN with dual dilated layers that combine dilated convolutions with large and small dilation factors. Global2Local~\cite{gao2021global2local} proposes a global-local search scheme to search for effective receptive field combinations instead of hand-designed patterns. C2F-TCN~\cite{singhania2021coarse} combines U-Net-like architecture and temporal convolutions, resulting in a coarse-to-fine structure. ASFormer~\cite{ASFormer} firstly introduces attention modules to action segmentation. In this paper, Transformer is integrated into the U-Net-like architecture, resulting in a pure Transformer model without temporal convolutions.
\\
{\bf Transformer} \quad 
Transformer~\cite{Transformer} was originally designed for sequence-to-sequence tasks in NLP. Afterward, Transformer-based models have succeeded in many other fields because of their powerful modeling capabilities and flexible architecture. For example, many researchers apply hybrid Transformer models in different video understanding tasks including visual tracking~\cite{yan2021learning}, video instance segmentation~\cite{wang2021end} and video action recognition~\cite{girdhar2019video}. To our best knowledge, the only Transformer-based model for the action segmentation task is ASFormer~\cite{ASFormer}. However, ASFormer utilizes 1D dilated convolution in each layer to bring in strong inductive priors, which does not break through the limitations of convolution. In this paper, we propose a pure Transformer model which combines the U-Net-like architecture and Transformer. Besides, Informer~\cite{informer} selects dominant queries by measuring the distance between their attention probability distributions and uniform distribution to reduce the complexity of self-attention. Anomaly Transformer~\cite{xu2021anomaly} utilizes discrepancy between each time point’s prior-association and its series-association to detect anomalies. Differently, we minimize the distance between the prior distribution for boundary frames and the distribution of local attentions to enhance the ability to discriminate boundaries.

\section{Preliminary}
In this section, we recap the preliminaries in Transformer and the problem formulation of action segmentation.

{\bf Transformer} \quad
Each layer in Transformer consists of two main components: a multi-head self-attention module and a feed-forward network ($\textup{FFN}$). We denote the input of the self-attention module as $H\in \mathbb{R}^{T \times d}$, where $T$ and $d$ are the length and dimension of the input, respectively. For the single-head situation, the input is projected by three matrices $W_Q\in \mathbb{R}^{d\times d_Q}$, $W_K\in \mathbb{R}^{d\times d_K}$ and $W_V\in \mathbb{R}^{d \times d_V}$ to obtain the query, key and value matrices: $Q=\{q_1,\ldots,q_T\}$, $K=\{k_1,\ldots,k_T\}$ and $V=\{v_1,\ldots,v_T\}$. Then the attention calculation is given by:
\begin{small}
\begin{equation}
\label{eq:fullattention}
\textup{Attn}(Q,K,V) = \textup{Softmax}(\frac{QK^T}{\sqrt{d_K}})V=\textup{Softmax}(A)V 
\end{equation}
\end{small}
where $A$ is the \textit{attention matrix} consisting of all the similarity scores between any query-key pair.

{\bf Action Segmentation} \quad
Given an input long video, we first extract a feature vector from each frame, so that the video is represented by a sequence of frame-wise features: \({x} = [{x}_1, \cdots, {x}_T] \in \mathbb{R}^{T \times d}\), where \({x}_t\) is the feature of the \(t\)-th frame, \(d\) is the dimensionality of \({x}_t\), and \(T\) is the length of the video. Our goal is to predict the class label \(\hat{c}_t \in \left\{1,\cdots,C \right\}\) for each frame \({x}_t\), resulting in the prediction \([\hat{c}_1, \cdots, \hat{c}_T]\), where \(C\) is the number of classes.

\section{Methodology}

\subsection{Temporal U-Transformer Architecture}
In this section, we present our model in detail. TUT contains a prediction generation stage and $M$ refinement stages following the multi-stage architecture in MS-TCN. The generation stage generates initial segmentation predictions while each refinement stage refines the predictions of the previous stage. These stages have the same architecture. As shown in Fig.~\ref{fig:model}, each stage can be separated into four components: the input projection, the encoder consisting of $N$ identical encoder layers, the decoder consisting of $N$ identical decoder layers, and the output classifier. Both the input projector and the output classifier are fully connected layers, which reduce the input dimension to feed the encoder and classify the output of the decoder, respectively.

\textbf{Local Attention}\quad
In the original self-attention module of Transformer, any query $q_i$ needs to calculate the similarity scores with all the keys $\{k_1,\ldots,k_T\}$ to generate the attention matrix $A$, which leads to quadratic complexity, i.e., the complexity is $\mathcal{O}(T^2)$. Restricting the attention computation to a local window with a fixed size $w$ can reduce the operation to linear complexity $\mathcal{O}(wT)$, which is called \textit{local attention}~\cite{beltagy2020longformer}. At this point, each query $q_i$ only needs to calculate the similarity with those keys in the window centered on its position, i.e., $\{k_{s},\ldots,k_i,\ldots,k_{e}\}$, where $s=\max \{i-\lfloor \frac{w}{2} \rfloor, 0\}$ and $e=\min \{i+\lfloor \frac{w}{2} \rfloor, T\}$ represent the start and end position respectively. Therefore, the output of the $i$-th position is:

\begin{equation}
o_i^T = \textup{Softmax}(\frac{q_i^T k_s}{\sqrt{d_k}},\ldots,\frac{q_i^T k_e}{\sqrt{d_k}})(v_s,\ldots,v_e)^T,
\end{equation}
\begin{equation}
\label{eq:localattention}
\textup{LA}(Q,K,V) = [o_1, o_2, \ldots, o_T].
\end{equation}

The local attention ($\textup{LA}$) does not narrow down the overall receptive field of the model. Due to temporal sampling between layers, the receptive field increases exponentially with the number of layers, which is sufficient to cover the entire video sequence to capture global and local dependencies. 

\textbf{Scale-Shared Positional Encoding}\quad
Since the attention mechanism in Transformer can not perceive the positions of elements, many works \cite{dai2019transformer, ke2020rethinking} adopt various strategies to introduce positional information. Since the lengths of untrimmed videos in action segmentation tasks are usually too long and vary drastically, absolute position encoding will influence the performance. Therefore, we employ the learnable relative positional encoding \cite{shaw2018self}, the basic idea of which is to embed the relative distances of all query-key pairs as scalars and add them to the attention matrix.

Considering the distance between any two frames within a local window does not exceed the window size $w$, we can get the relative position encoding $R_{ij}$ between $q_i$ and $k_j$ by a learnable embedding matrix $W_{rpe}\in \mathbb{R}^{w \times h}$, where $i,j$ represent the position index and $h$ is the number of heads. The resulting positional encoding $R$ will be added to the corresponding positions of the attention matrices in different heads. Layers with the same layer index in different stages process inputs with the same temporal resolution, and their RPEs should be the same. Therefore, we adopt a scale-shared strategy, i.e., corresponding layers with the same scale in different stages share the same $W_{rpe}$. 

\textbf{Fine-to-Abstract Encoder}\quad 
The encoder is composed of $N$ identical encoder layers. As shown in Fig.~\ref{fig:model}, it is similar to the encoder in the vanilla Transformer but there are three differences. Firstly, there exists a nearest-neighbor downsampling process at the beginning of each layer, which halves the input temporal dimension. Secondly, the full attention is replaced by the local attention with scale-shared relative position encoding. Thirdly, we utilize the instance normalization ($\textup{IN}$)~\cite{ulyanov2016instance} instead of the layer normalization~\cite{layernorm}. In summary, the set of operations at the $l$-th encoder layer can be formally described as follows: 
\begin{footnotesize}
\begin{equation}
\begin{aligned}
H^{l,1}_{en} &= \textup{DownSample}(H^{l-1}_{en}),\\
H^{l,2}_{en} &= \textup{IN}(\textup{LA}(H^{l,1}_{en}W_{Q}^l, H^{l,1}_{en}W_{K}^l, H^{l,1}_{en}W_{V}^l)+H^{l,1}_{en}),\\
H^{l}_{en} &= \textup{IN}(\textup{FFN}(H^{l,2}_{en})+H^{l,2}_{en}).
\end{aligned}
\end{equation}
\end{footnotesize}

\textbf{Abstract-to-Fine Decoder}\quad 
The decoder consisting of $N$ identical layers is symmetric to the encoder. In each decoder layer, the temporal upsampling is utilized to gradually restore the original temporal resolution of input frames. The upsampling process is also implemented by nearest interpolation. We do not concatenate the encoder layer output and the previous layer input as the decoder layer input like the original U-Net~\cite{ron2015unet}, which will take up more memory. To keep the hidden dimension, we modify the cross-attention in the original Transformer to leverage the information from the encoder. Specifically, in our local cross attention, the query is generated by the output of the previous decoder layer, while the value and key both come from the output of the corresponding encoder layer having the same temporal dimension as the query. Therefore, the $l$-th decoder layer generates the representation $H_{de}^{l}$ from $H_{de}^{l-1}$ of the $(l\!-\!1)$-th layer as follows:
\begin{footnotesize}
\begin{equation}
\begin{aligned}
H^{l,1}_{de} &= \textup{UpSample}(H_{de}^{l-1}),\\
H^{l,2}_{de} &= \textup{IN}(\textup{LA}(H_{de}^{l,1}W_{Q}^l, H_{en}^{N-l}W_{K}^l, H_{en}^{N-l}W_{V}^l)+H_{de}^{l,1}),\\
H^{l}_{de} &= \textup{IN}(\textup{FFN}(H_{de}^{l,2})+H_{de}^{l,2}).
\end{aligned}
\end{equation}
\end{footnotesize}

\subsection{Boundary-aware Loss}
During the training phase, we combine three different losses: the frame-wise classification loss $\mathcal{L}_{CE}$, the smoothing loss $\mathcal{L}_{TMSE}$ in MS-TCN~\cite{farha2019ms}, and our proposed boundary-aware loss $\mathcal{L}_{BA}$. Since the loss function of each stage is exactly the same, we only analyze the loss of the $s$-th stage $\mathcal{L}^s$. 

Following~\cite{farha2019ms}, we use a cross-entropy loss as $\mathcal{L}_{CE}^s$, and a truncated mean squared error over the frame-wise log-probabilities as $\mathcal{L}_{TMSE}^s$ in the $s$-th stage:
\begin{small}
\begin{equation}
\label{eq:celoss}
\mathcal{L}_{CE}^s = \frac{1}{T} \sum\limits_{t = 1}^T -log({y}^s_t(c_t)), 
\end{equation}
\end{small}
\begin{small}
\begin{equation}
\label{eq:tmseloss} 
\mathcal{L}_{TMSE}^s = \frac{1}{TC} \sum\limits_{t = 1}^T \sum\limits_{c = 1}^C (max(\lvert log({y}^s_t(c)) - log({y}^s_{t-1}(c)) \rvert,\theta))^2,
\end{equation}
\end{small}
where \({y}^s_t(c_t)\) is the predicted probability that \({x}_t\) belongs to the \(c_t\)-th class, and \(\theta=4\) is a pre-set threshold. In $\mathcal{L}_{TMSE}^s$, gradients are not calculated w.r.t. \({y}^s_{t-1}(c)\).

\begin{figure}[t]
\centering
\includegraphics[width=0.45\textwidth]{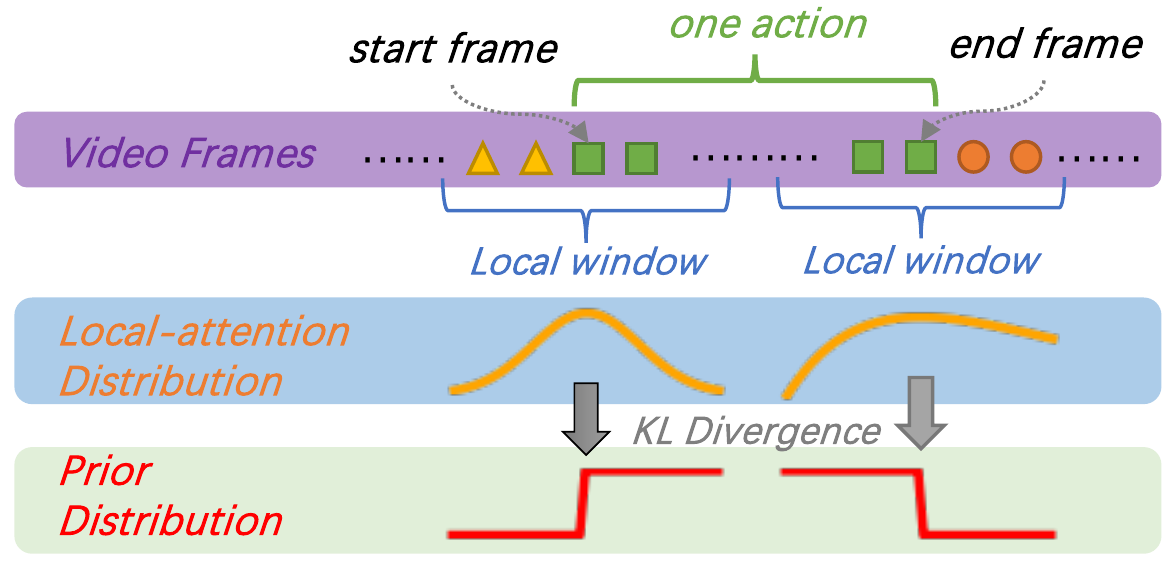} 
\caption{Schematic diagram of the boundary-aware loss. Frames of the same color belong to the same class. We can extract the local-attention distributions of the start and end frame from the attention matrix. The boundary-aware loss aims to minimize the KL divergence between the local-attention distribution and the prior distribution.}
\label{fig:BALoss}
\end{figure}

Action boundaries are vital for video action segmentation. Frame-wise classification treats boundary frames and intermediate frames equally, which causes the features learned to be not sensitive to boundaries. We propose a novel boundary-aware loss to enhance the ability to discriminate boundaries. The boundary-aware loss regularizes feature learning by imposing additional constraints on the attention matrix in the local attention module.

\textbf{Prior Distribution}\quad 
Intuitively, a frame is more likely to be a start frame if it has more similarities with the neighboring frames in its backward direction and less with those in its forward direction. Conversely, if a frame is similar to the neighboring frames in its forward direction but different from those in its backward direction, it is probably an end frame. Therefore, the similarity distribution between a boundary frame $i$ (anchor) and its neighbors should exhibit two different patterns, depending on whether the anchor is the start frame or the end frame. We use the adapted sign function as the two {\em prior distributions} corresponding to the above two patterns:
\begin{footnotesize}
\begin{equation}
\label{eq:prior}
\mathcal{P}_i =
\begin{cases}
\textup{Rescale}(\textup{Sgn}(j-i)+1) & i \in \{ start \, frames\}, \\
\textup{Rescale}(-\textup{Sgn}(j-i)+1) & i \in \{ end \, frames \}, \\
\end{cases}
\end{equation}
\end{footnotesize}
where $j\in [i-\lfloor \frac{w}{2} \rfloor, i+\lfloor \frac{w}{2} \rfloor ]$ is a frame within the local window, and $(j-i)$ means the distance between frame $j$ and anchor $i$. $\{ start \, frames\}$ and $\{ end \, frames \}$ represent the set of start frames and the set of end frames, respectively. $\textup{Sgn}(x)$ is 1 when $x$ is greater than or equal to 0 and -1 when $x$ is less than 0. Further, we use $\textup{Rescale}(\cdot)$ to transform the sum of probabilities to 1.

\textbf{Local-Attention Distribution}\quad 
The attention matrix in the local attention module consists of all the similarity scores between query-key pairs. We can extract the similarity distribution of the anchor and its neighboring frames from the attention matrix $A$, named as {\em local-attention distribution} of the anchor frame $i$:
\begin{footnotesize}
\begin{equation}
\label{eq:lad}
\mathcal{D}_i = A\left [i,(i-\lfloor \frac{w}{2} \rfloor):(i+\lfloor \frac{w}{2} \rfloor)\right ].
\end{equation}
\end{footnotesize}
As shown in Fig.~\ref{fig:BALoss}, we introduce a boundary-aware loss to approximate the local-attention distribution of the boundary to the corresponding prior distribution, which can be implemented by minimizing the symmetrized KL divergence between two distributions of the $t$-th frame:
\begin{equation}
\label{eq:baloss} 
\mathcal{L}_{BA}^s = \frac{1}{T} \sum\nolimits_{t=\lfloor \frac{w}{2} \rfloor}^{T-\lfloor \frac{w}{2} \rfloor} \textup{KL}(\mathcal{P}_t^s||\mathcal{D}_t^s).
\end{equation}
We can freely obtain boundary labels from class labels and the similarity distribution of each boundary from the local attention module. Therefore, we can calculate the boundary-aware loss without additional modules and annotations. We only compute it in the first layer of the encoder and the last layer of the decoder since temporal downsampling blurs high-level boundaries. Therefore, the final loss for the $s$-th stage is the weighted sum of the three losses:
\begin{equation}
\label{eq:stageloss} 
\mathcal{L}^s = \mathcal{L}_{CE}^s + \lambda \mathcal{L}_{TMSE}^s + \beta \mathcal{L}_{BA}^s.
\end{equation}
We set \(\lambda = 0.15\) and $\beta$ is an adjustable hyper-parameter.

The overall loss function of all stages in the training phase is $\mathcal{L} = \sum_s\mathcal{L}^s$.
In the testing phase, we use the frame classification results of the last refinement stage as our segmentation results.



\section{Experiments}
\label{sec:exp}
\subsection{Datasets}
We empirically perform experiments on three public benchmark datasets: the 50Salads dataset~\cite{stein2013combining}, the Georgia Tech Egocentric Activities dataset~\cite{fathi2011learning}, and the Breakfast dataset~\cite{kuehne2014language}. More details about these datasets are given in the appendix.

\textbf{GTEA}: It consists of 28 egocentric videos with 11 action classes including the background class from 7 activities. On average, there are about 20 action instances in each video.

\textbf{50Salads}: It consists of 50 top-view videos with 17 action classes. On average, each video lasts for about 6.4 minutes and contains about 20 action instances.

\textbf{Breakfast}: It consists of 1,712 third-person view videos with 48 action classes. On average, each video contains about 6 action instances.

On all datasets, we represent each video as a sequence of visual features. We employ the I3D~\cite{carreira2017quo} frame-wise features provided in MS-TCN++~\cite{li2020ms} as inputs. The temporal video resolution is fixed to 15 fps on all datasets. Following MS-TCN++~\cite{li2020ms}, we also use the segmental F1 score at overlapping thresholds 10\%, 25\%, 50\% (F1@{10,25,50}), segmental edit distance (Edit) measuring the difference between predicted segment paths and ground-truth segment paths, and frame-wise accuracy (Acc) as evaluation metrics. Following the general settings in action segmentation~\cite{farha2019ms,li2020ms}, we perform five-fold cross-validation on the 50Salads dataset and four-fold cross-validation on the GTEA and Breakfast datasets and report the average performances.

\subsection{Experimental Details}
We briefly summarize the basics, and more information and setups are given in the appendix.

\begin{table*}[t]
\centering
\small
\begin{tabular}{l|ccccc|ccccc|ccccc}
\toprule
Dataset & \multicolumn{5}{c|}{50Salads} & \multicolumn{5}{c|}{GTEA} & \multicolumn{5}{c}{Breakfast} \\
\cmidrule{2-16}   Metric & \multicolumn{3}{c}{F1@\{10,25,50\}} & Edit & Acc & \multicolumn{3}{c}{F1@\{10,25,50\}} & Edit & Acc & \multicolumn{3}{c}{F1@\{10,25,50\}} & Edit & Acc \\
\midrule
IDT+LM      & 44.4 & 38.9 & 27.8 & 45.8 & 48.7 & - & - &  - & - & - & - & - & - & - & -\\
Bi-LSTM     & 62.6 & 58.3 & 47.0 & 55.6 & 55.7 & 66.5 & 59.0 & 43.6 & - & 55.5 & - & - & - & - & -\\
Dilated TCN & 52.2 & 47.6 & 37.4 & 43.1 & 59.3 & 58.8 & 52.2 & 42.2 & - & 58.3 & - & - & - & - & -\\
ST-CNN      & 55.9 & 49.6 & 37.1 & 45.9 & 59.4 & 58.7 & 54.4 & 41.9 & - & 60.6 & - & - & - & - & -\\
ED-TCN      & 68.0 & 63.9 & 52.6 & 52.6 & 64.7 & 72.2 & 69.3 & 56.0 & - & 64.0 & - & - & - & - & 43.3\\
TDRN        & 72.9 & 68.5 & 57.2 & 66.0 & 68.1 & 79.2 & 74.4 & 62.7 & 74.1 & 70.1 & - & - & - & - & -\\
MS-TCN      & 76.3 & 74.0 & 64.5 & 67.9 & 80.7 & 85.8 & 83.4 & 69.8 & 79.0 & 76.3 & 52.6 & 48.1 & 37.9 & 61.7 & 66.3\\
MS-TCN++    & 80.7 & 78.5 & 70.1 & 74.3 & 83.7 & 88.8 & 85.7 & 76.0 & 83.5 & \underline{80.1} & 64.1 & 58.6 & 45.9 & 65.6 & 67.6\\
BCN         & 82.3 & 81.3 & 74.0 & 74.3 & 84.4 & 88.5 & 87.1 & 77.3 & 84.4 & 79.8 & 68.7 & 65.5 & 55.0 & 66.2 & 70.4\\
Global2Local& 80.3 & 78.0 & 69.8 & 73.4 & 82.2 & 89.9 & 87.3 & 75.8 & \underline{84.6} & 78.5 & 74.9 & 69.0 & 55.2 & 73.3 & 70.7\\
ASRF        & 84.9 & 83.5 & 77.3 & 79.3 & 84.5 & 89.4 & \underline{87.8} & \textbf{79.8} & 83.7 & 77.3 & 74.3 & 68.9 & 56.1 & 72.4 & 67.6\\
C2F-TCN     & 84.3 & 81.8 & 72.6 & 76.4 & 84.9 & \textbf{90.3} & \textbf{88.8} & 77.7 & \textbf{86.4} & \textbf{80.8} & 72.2 & 68.7 & 57.6 & 69.6 & \textbf{76.0} \\
ASFormer    & 85.1 & 83.4 & 76.0 & 79.6 & 85.6 & \underline{90.1} & \textbf{88.8} & \underline{79.2} & \underline{84.6} & 79.7 & \underline{76.0} & 70.6 & 57.4 & \textbf{75.0} & 73.5\\
\midrule
TUT$^\dag$  & \underline{87.7} & \underline{87.1} & \underline{79.9} & \underline{82.6} & \underline{85.9} & 88.1 & 86.2 & 71.2 & 83.2 &  76.4 &  \underline{76.0} & \underline{71.7} & \underline{59.5} & \underline{73.7} & \underline{75.5}\\
TUT & \textbf{89.3} & \textbf{88.3} & \textbf{81.7} & \textbf{84.0} & \textbf{87.2} & 89.0 & 86.4 & 73.3 & 84.1 & 76.1 & \textbf{76.2} & \textbf{71.9} & \textbf{60.0} & \underline{73.7} & \textbf{76.0}\\
\bottomrule
\end{tabular}
\caption{Comparison with the state-of-the-art results on three datasets. TUT$^\dag$ is the proposed TUT model trained without the boundary-aware loss. \textbf{Bold} and \underline{underlined} denote the best and second-best results in each column, respectively.}
\label{tab:sota}
\end{table*}

\begin{figure*}[t]
\centering
\includegraphics[width=1\textwidth]{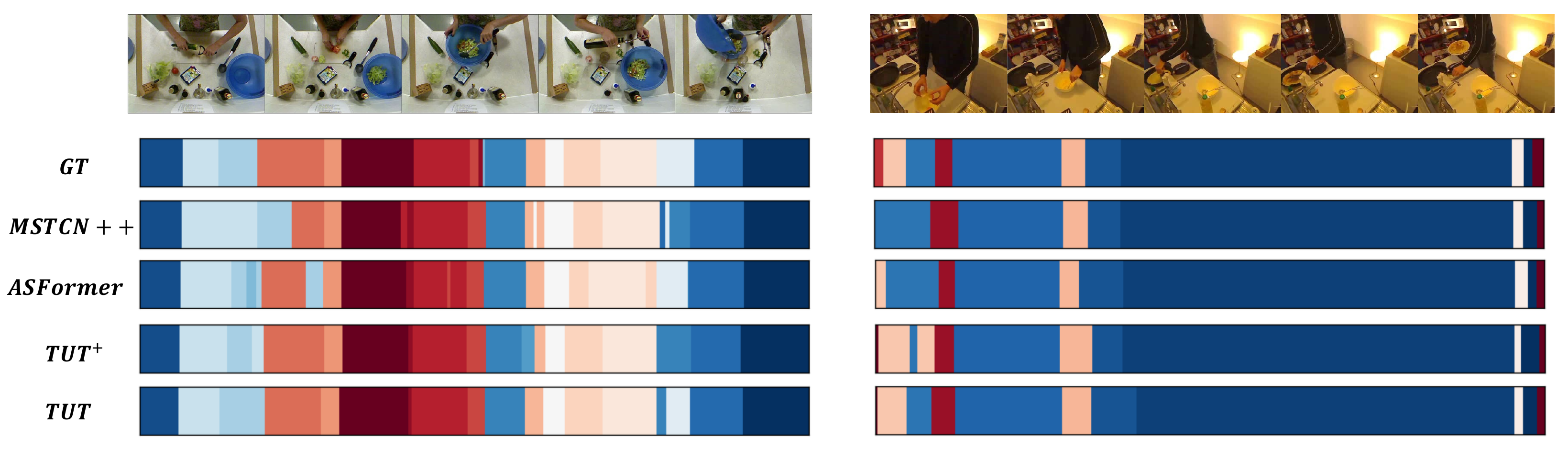} 
\caption{Visualization of segmentation results on (left) 50Salads and (right) Breakfast datasets. The same color represents the same action class. Our TUT model has more accurate action classification results compared to ASFormer and MS-TCN++, and the TUT model trained with boundary-aware loss can further reduce over-segmentation errors.}
\label{fig:result}
\end{figure*}

\textbf{Baselines}\quad
We compare our proposed model TUT with various previous state-of-the-art action segmentation methods including RNN-based and TCN-based models. The compared methods include IDT+LM~\cite{richard2016temporal}, Bi-LSTM~\cite{singh2016multi}, Dilated TCN~\cite{lea2017temporal}, ST-CNN~\cite{lea2016segmental}, ED-TCN~\cite{lea2017temporal}, TDRN~\cite{lei2018temporal}, MS-TCN~\cite{farha2019ms}, MS-TCN++~\cite{li2020ms}, BCN~\cite{BCN}, Global2Local~\cite{gao2021global2local}, ASRF~\cite{asrf}, C2F-TCN~\cite{singhania2021coarse}, and ASFormer~\cite{ASFormer}. 

\textbf{Implementation details}\quad
We employ the ADAM optimizer~\cite{adam}  for a maximum number of 150 epochs to train our model on all datasets. The batch size is set to 1. All experiments are implemented with PyTorch~\cite{pytorch} and conducted on one NVIDIA TITAN RTX 24GB GPU. All ablation studies about hyper-parameters and model structures are performed on the 50Salads dataset.

\subsection{Comparison and Result Analysis}

In Table~\ref{tab:sota}, we compare our proposed model with the state-of-the-art methods on three datasets. For a fair comparison of all models, we list the results of training our model without the additional boundary-aware loss, which corresponds to TUT$^\dag$. To demonstrate the effectiveness of our proposed boundary-aware loss, we also report the results of our model jointly trained with the additional boundary-aware loss, called TUT.

On the smallest dataset, i.e., GTEA, our performance is slightly behind the top methods, e.g. C2F-TCN. This is reasonable since GTEA has only 21 training samples per split, which are not adequate to train our pure Transformer model. But our model is still better than some other models such as MS-TCN, which also proves the superiority of the attention mechanism. The 50Salads dataset has the longest average video length, while the Breakfast dataset has the largest number of video samples. On these larger datasets, TUT significantly outperforms the previous TCN-based backbone models on all metrics by a large margin, which proves the advantages of Transformer-based models in processing large-scale sequence data. Even though the state-of-the-art TCN-Based model (C2F-TCN) utilizes data augmentation and additional video-level action loss that we do not employ, TUT$^\dag$ outperforms it on these two datasets. On the 50Salads dataset, our backbone TUT$^\dag$ beats all previous models on all metrics. Regardless of whether our model is trained with or without the boundary-aware loss, the results of our model are better than ASFormer. Specifically, compared to ASFormer, TUT$^\dag$ gives \textbf{2.6}\% (85.1$\to$87.7), \textbf{3.7}\% (83.4$\to$87.1), \textbf{3.9}\% (76.0$\to$79.9) improvements on three F1 scores, \textbf{3.0}\% (79.6$\to$ 82.6) improvement on the segmental edit distance, and \textbf{0.3}\% (85.6$\to$85.9) improvement on the accuracy. On the Breakfast dataset, our model achieves the best performance on the F1 metrics, while the performance on Edit and Acc ranks in the top two of all methods. We also visualize some prediction results of our model in Fig.~\ref{fig:result}. 

Both ASRF and BCN use TCN-based models as the feature enhancement backbone. Similar to our boundary-aware loss, they utilize additional auxiliary tasks and training loss to improve performance, which can also be used in our Transformer-based models to expect further performance improvements. Besides, TUT outperforms TUT$^\dag$ on all the datasets, which demonstrates the effectiveness of the proposed boundary-aware loss. Models trained with the boundary-aware loss are better able to recognize boundary frames, which leads to improvements in all metrics.

{\bf Effect of the boundary-aware loss}\quad
The quantitative impact of the boundary-aware loss on performance can already be observed in Table~\ref{tab:sota}. As shown in Fig.~\ref{fig:result}, we observe that TUT$^\dag$ trained without the boundary-aware loss does not accurately locate the boundaries of some action segments and sometimes incorrectly misclassifies intermediate frames as boundary frames, resulting in over-segmentation. The TUT jointly trained with the boundary-aware loss largely improves these problems. Therefore, both the qualitative results in Fig.~\ref{fig:result} and the quantitative metrics in Table~\ref{tab:sota} demonstrate the effectiveness of boundary-aware loss. Further ablation experiments on boundary-aware loss, e.g., the weight hyper-parameter $\beta$, and other metrics to measure the distribution distance, can be found in the appendix.

\begin{table}[t]
\centering
\small
\begin{tabular}{ccccccc}
\toprule
PE & Strategy & \multicolumn{3}{c}{F1@\{10,25,50\}} & Edit  & \multicolumn{1}{c}{Acc} \\
\midrule
\ding{56} & \ding{56} & 86.4 & 85.4 & 78.6 & 80.5 & 85.4 \\
\midrule  
\multirow{2}[4]{*}{APE} & Sinusoidal & 82.8 & 80.6 & 71.3 & 77.0 & 82.0\\
\cmidrule{2-2}          & Learnable & 80.5 & 77.7 & 69.1 & 73.5 & 82.0\\
\midrule    
\multirow{3}[6]{*}{RPE} & No Share & 87.1 & 86.6 & 79.4 & 81.9 & \textbf{86.0} \\
\cmidrule{2-2}    & Stage-Shared & 85.8 & 85.3 & 78.6 & 80.9 & 85.4 \\
\cmidrule{2-2}    & Scale-Shared & \textbf{87.7} & \textbf{87.1} & \textbf{79.9} & \textbf{82.6} & \textbf{85.9} \\
\bottomrule
\end{tabular}
\caption{Comparison of different positional encoding. \textit{APE} and \textit{RPE} refer to the absolute positional encoding and relative positional encoding, respectively. We also explore three ways of applying RPE: all layers do not share $W_{rpe}$ (\textit{No Share}), all layers in each stage share $W_{rpe}$ (\textit{Stage-Shared}), and corresponding layers with the same scale in different stages share $W_{rpe}$ (\textit{Scale-Shared}).}
\label{tab:rpe}
\end{table}

{\bf Comparison of different positional encoding}\quad
To verify the effectiveness of relative positional encoding (RPE) in TUT, we compare different positional encoding (PE) methods in Table~\ref{tab:rpe}. Since the lengths of video samples vary over a large span, we observe that inflexible absolute positional encoding (APE) leads to performance degradation. The overall performance of RPE is better than that without PE, which illustrates the importance of position information. We also compare three different sharing ways of applying RPE. Stage-Shared has the worst performance since it applies the same RPE to layers that deal with inputs of different resolutions. Therefore, layers of different scales can not share RPE. Scale-Shared performs best, which means that the attention patterns of layers with the same scale are similar, even if they belong to different stages.  

\begin{table}[t]
\centering
\resizebox{1\columnwidth}{!}{
\small
\begin{tabular}{cccccccc}
\toprule
Architure & Attention & \multicolumn{3}{c}{F1@\{10,25,50\}} & Edit  & Acc & GPU Mem. \\
\midrule
\multirow{3}[5]{*}{Standard} & Full  & 4.6 & 2.8 & 1.4 & 3.3 & 62.8 & 18.7G\\
\cmidrule{2-2}          & LogSparse & 56.2 & 51.7 & 41.2 & 45.3 & 69.0 & 18.7G\\
\cmidrule{2-2}          & Local & 74.6 & 72.2 & 63.1 & 64.8 & 81.0 & 4.6G\\
\midrule     
\multirow{3}[4]{*}{U-Trans} & Full  & 35.1 & 25.4 & 9.8 & 31.9 & 43.8 &  9.9G\\
\cmidrule{2-2}          & LogSparse & 73.3 & 71.9 & 63.7 & 65.1 & 80.3 & 9.9G\\
\cmidrule{2-2}          & Local & \textbf{86.5} & \textbf{85.3} & \textbf{76.9} & \textbf{80.6} & \textbf{84.4} & 2.8G\\
\bottomrule
\end{tabular}}
\caption{Comparison of different architectures and attention patterns. \textit{Standard} refers to architecture without temporal sampling, where the temporal resolution of each layer is unchanged. And \textit{U-Trans} refers to our proposed architecture. \textit{Full}, \textit{LogSparse}, and \textit{Local} are three attention patterns.}
\label{tab:model}
\end{table}

{\bf Ablations of the architecture and attention patterns}\quad
We compare two model architectures and three attention patterns, resulting in a total of six combinations in Table~\ref{tab:model}. We remove all the temporal sampling in TUT to get the standard architecture. Attention patterns include full attention in the original Transformer~\cite{Transformer}, local attention, and LogSparse attention proposed in~\cite{ConvTransformer}, where each cell only attends to those cells whose distance from it increases exponentially. The reason we compare LogSparse is that its attention pattern is somewhat similar to the pattern in MS-TCN where the convolutional dilation factor exponentially increases with layers. In the standard architecture, long video inputs will cause out-of-memory. For a fair comparison, we downsample video samples below 5000 frames and control all model configurations and training parameters to be consistent. Considering that RPE introduces additional parameters, we do not use RPE in these ablation experiments. The U-Transformer architecture achieves better performance than the standard architecture with less GPU memory consumption. Full attention fails regardless of the architecture, showing that training on small data requires sparser attention patterns. Since adjacent frames usually have a stronger correlation in action segmentation, local attention performs much better than LogSparse attention.

More ablation studies about hyper-parameters are presented in the appendix.


\section{Conclusion}
In action segmentation tasks, most popular deep learning methods use temporal convolutional networks as their backbones. However, temporal convolutions limit the performance of these methods. For the first time, we propose a model without temporal convolutions which combines the temporal sampling and Transformer to construct a temporal U-Transformer architecture. The temporal downsampling and local attention modules together enable our model to very long videos. Furthermore, we propose a novel boundary-aware loss based on the local-attention distributions of boundary frames, which serves as a regularization term to train the model and can further enhance the ability of discriminating boundaries.

\appendix

\bibliography{aaai23}

\begin{thebibliography}{52}
\providecommand{\natexlab}[1]{#1}

\bibitem[{Ba, Kiros, and Hinton(2016)}]{layernorm}
Ba, J.~L.; Kiros, J.~R.; and Hinton, G.~E. 2016.
\newblock Layer normalization.
\newblock \emph{arXiv preprint arXiv:1607.06450}.

\bibitem[{Bai, Kolter, and Koltun(2018)}]{TCN}
Bai, S.; Kolter, J.~Z.; and Koltun, V. 2018.
\newblock An empirical evaluation of generic convolutional and recurrent
  networks for sequence modeling.
\newblock \emph{arXiv preprint arXiv:1803.01271}.

\bibitem[{Beltagy, Peters, and Cohan(2020)}]{beltagy2020longformer}
Beltagy, I.; Peters, M.~E.; and Cohan, A. 2020.
\newblock Longformer: The long-document transformer.
\newblock \emph{arXiv preprint arXiv:2004.05150}.

\bibitem[{Bhattacharya et~al.(2014)Bhattacharya, Kalayeh, Sukthankar, and
  Shah}]{bhattacharya2014recognition}
Bhattacharya, S.; Kalayeh, M.~M.; Sukthankar, R.; and Shah, M. 2014.
\newblock Recognition of complex events: Exploiting temporal dynamics between
  underlying concepts.
\newblock In \emph{Proceedings of the IEEE conference on computer vision and
  pattern recognition}, 2235--2242.

\bibitem[{Cao et~al.(2021)Cao, Wang, Chen, Jiang, Zhang, Tian, and
  Wang}]{cao2021swinunet}
Cao, H.; Wang, Y.; Chen, J.; Jiang, D.; Zhang, X.; Tian, Q.; and Wang, M. 2021.
\newblock Swin-unet: Unet-like pure transformer for medical image segmentation.
\newblock \emph{arXiv preprint arXiv:2105.05537}.

\bibitem[{Carreira and Zisserman(2017)}]{carreira2017quo}
Carreira, J.; and Zisserman, A. 2017.
\newblock Quo vadis, action recognition? a new model and the kinetics dataset.
\newblock In \emph{proceedings of the IEEE Conference on Computer Vision and
  Pattern Recognition}, 6299--6308.

\bibitem[{Cheng et~al.(2014)Cheng, Fan, Pankanti, and
  Choudhary}]{cheng2014temporal}
Cheng, Y.; Fan, Q.; Pankanti, S.; and Choudhary, A. 2014.
\newblock Temporal sequence modeling for video event detection.
\newblock In \emph{Proceedings of the IEEE Conference on Computer Vision and
  Pattern Recognition}, 2227--2234.

\bibitem[{Connor, Atlas, and Martin(1992)}]{rnn}
Connor, J.; Atlas, L.~E.; and Martin, D.~R. 1992.
\newblock Recurrent networks and NARMA modeling.
\newblock In \emph{In Advances in Neural Information Processing Systems},
  301--308.

\bibitem[{Dai et~al.(2019)Dai, Yang, Yang, Carbonell, Le, and
  Salakhutdinov}]{dai2019transformer}
Dai, Z.; Yang, Z.; Yang, Y.; Carbonell, J.~G.; Le, Q.~V.; and Salakhutdinov, R.
  2019.
\newblock Transformer-XL: Attentive Language Models beyond a Fixed-Length
  Context.
\newblock In \emph{ACL (1)}.

\bibitem[{Deng et~al.(2009)Deng, Dong, Socher, Li, Li, and
  Fei-Fei}]{deng2009imagenet}
Deng, J.; Dong, W.; Socher, R.; Li, L.-J.; Li, K.; and Fei-Fei, L. 2009.
\newblock Imagenet: A large-scale hierarchical image database.
\newblock In \emph{2009 IEEE conference on computer vision and pattern
  recognition}, 248--255. Ieee.

\bibitem[{Dosovitskiy et~al.(2020)Dosovitskiy, Beyer, Kolesnikov, Weissenborn,
  Zhai, Unterthiner, Dehghani, Minderer, Heigold, Gelly
  et~al.}]{dosovitskiy2020vit}
Dosovitskiy, A.; Beyer, L.; Kolesnikov, A.; Weissenborn, D.; Zhai, X.;
  Unterthiner, T.; Dehghani, M.; Minderer, M.; Heigold, G.; Gelly, S.; et~al.
  2020.
\newblock An Image is Worth 16x16 Words: Transformers for Image Recognition at
  Scale.
\newblock In \emph{International Conference on Learning Representations}.

\bibitem[{Farha and Gall(2019)}]{farha2019ms}
Farha, Y.~A.; and Gall, J. 2019.
\newblock Ms-tcn: Multi-stage temporal convolutional network for action
  segmentation.
\newblock In \emph{Proceedings of the IEEE Conference on Computer Vision and
  Pattern Recognition}, 3575--3584.

\bibitem[{Fathi, Ren, and Rehg(2011)}]{fathi2011learning}
Fathi, A.; Ren, X.; and Rehg, J.~M. 2011.
\newblock Learning to recognize objects in egocentric activities.
\newblock In \emph{CVPR 2011}, 3281--3288. IEEE.

\bibitem[{Gao et~al.(2021)Gao, Han, Li, Peng, Wang, and
  Cheng}]{gao2021global2local}
Gao, S.-H.; Han, Q.; Li, Z.-Y.; Peng, P.; Wang, L.; and Cheng, M.-M. 2021.
\newblock Global2local: Efficient structure search for video action
  segmentation.
\newblock In \emph{Proceedings of the IEEE/CVF Conference on Computer Vision
  and Pattern Recognition}, 16805--16814.

\bibitem[{Girdhar et~al.(2019)Girdhar, Carreira, Doersch, and
  Zisserman}]{girdhar2019video}
Girdhar, R.; Carreira, J.; Doersch, C.; and Zisserman, A. 2019.
\newblock Video action transformer network.
\newblock In \emph{Proceedings of the IEEE Conference on Computer Vision and
  Pattern Recognition}, 244--253.

\bibitem[{Huang, Sugano, and Sato(2020)}]{huang2020improving}
Huang, Y.; Sugano, Y.; and Sato, Y. 2020.
\newblock Improving Action Segmentation via Graph-Based Temporal Reasoning.
\newblock In \emph{Proceedings of the IEEE/CVF Conference on Computer Vision
  and Pattern Recognition}, 14024--14034.

\bibitem[{Ishikawa et~al.(2021)Ishikawa, Kasai, Aoki, and Kataoka}]{asrf}
Ishikawa, Y.; Kasai, S.; Aoki, Y.; and Kataoka, H. 2021.
\newblock Alleviating over-segmentation errors by detecting action boundaries.
\newblock In \emph{Proceedings of the IEEE/CVF Winter Conference on
  Applications of Computer Vision}, 2322--2331.

\bibitem[{Ji et~al.(2021)Ji, Zhang, Wang, Li, Wu, Zhang, and Luo}]{ji2021multi}
Ji, Y.; Zhang, R.; Wang, H.; Li, Z.; Wu, L.; Zhang, S.; and Luo, P. 2021.
\newblock Multi-compound transformer for accurate biomedical image
  segmentation.
\newblock In \emph{International Conference on Medical Image Computing and
  Computer-Assisted Intervention}, 326--336. Springer.

\bibitem[{Karaman, Seidenari, and Del~Bimbo(2014)}]{karaman2014fast}
Karaman, S.; Seidenari, L.; and Del~Bimbo, A. 2014.
\newblock Fast saliency based pooling of fisher encoded dense trajectories.
\newblock In \emph{ECCV THUMOS Workshop}, 5.

\bibitem[{Kay et~al.(2017)Kay, Carreira, Simonyan, Zhang, Hillier,
  Vijayanarasimhan, Viola, Green, Back, Natsev et~al.}]{kay2017kinetics}
Kay, W.; Carreira, J.; Simonyan, K.; Zhang, B.; Hillier, C.; Vijayanarasimhan,
  S.; Viola, F.; Green, T.; Back, T.; Natsev, P.; et~al. 2017.
\newblock The kinetics human action video dataset.
\newblock \emph{arXiv preprint arXiv:1705.06950}.

\bibitem[{Ke, He, and Liu(2020)}]{ke2020rethinking}
Ke, G.; He, D.; and Liu, T.-Y. 2020.
\newblock Rethinking Positional Encoding in Language Pre-training.
\newblock In \emph{International Conference on Learning Representations}.

\bibitem[{Kenton and Toutanova(2019)}]{kenton2019bert}
Kenton, J. D. M.-W.~C.; and Toutanova, L.~K. 2019.
\newblock BERT: Pre-training of Deep Bidirectional Transformers for Language
  Understanding.
\newblock In \emph{Proceedings of NAACL-HLT}, 4171--4186.

\bibitem[{Kingma and Ba(2014)}]{adam}
Kingma, D.~P.; and Ba, J. 2014.
\newblock Adam: A method for stochastic optimization.
\newblock In \emph{International Conference on Learning Representations}.

\bibitem[{Kuehne, Arslan, and Serre(2014)}]{kuehne2014language}
Kuehne, H.; Arslan, A.; and Serre, T. 2014.
\newblock The language of actions: Recovering the syntax and semantics of
  goal-directed human activities.
\newblock In \emph{Proceedings of the IEEE conference on computer vision and
  pattern recognition}, 780--787.

\bibitem[{Kuehne, Gall, and Serre(2016)}]{kuehne2016end}
Kuehne, H.; Gall, J.; and Serre, T. 2016.
\newblock An end-to-end generative framework for video segmentation and
  recognition.
\newblock In \emph{2016 IEEE Winter Conference on Applications of Computer
  Vision (WACV)}, 1--8. IEEE.

\bibitem[{Kuehne, Richard, and Gall(2017)}]{kuehne2017weakly}
Kuehne, H.; Richard, A.; and Gall, J. 2017.
\newblock Weakly supervised learning of actions from transcripts.
\newblock \emph{Computer Vision and Image Understanding}, 163: 78--89.

\bibitem[{Lea et~al.(2017)Lea, Flynn, Vidal, Reiter, and
  Hager}]{lea2017temporal}
Lea, C.; Flynn, M.~D.; Vidal, R.; Reiter, A.; and Hager, G.~D. 2017.
\newblock Temporal convolutional networks for action segmentation and
  detection.
\newblock In \emph{proceedings of the IEEE Conference on Computer Vision and
  Pattern Recognition}, 156--165.

\bibitem[{Lea et~al.(2016)Lea, Reiter, Vidal, and Hager}]{lea2016segmental}
Lea, C.; Reiter, A.; Vidal, R.; and Hager, G.~D. 2016.
\newblock Segmental spatiotemporal cnns for fine-grained action segmentation.
\newblock In \emph{European Conference on Computer Vision}, 36--52. Springer.

\bibitem[{Lei and Todorovic(2018)}]{lei2018temporal}
Lei, P.; and Todorovic, S. 2018.
\newblock Temporal deformable residual networks for action segmentation in
  videos.
\newblock In \emph{Proceedings of the IEEE Conference on Computer Vision and
  Pattern Recognition}, 6742--6751.

\bibitem[{Li et~al.(2019)Li, Jin, Xuan, Zhou, Chen, Wang, and
  Yan}]{ConvTransformer}
Li, S.; Jin, X.; Xuan, Y.; Zhou, X.; Chen, W.; Wang, Y.-X.; and Yan, X. 2019.
\newblock Enhancing the Locality and Breaking the Memory Bottleneck of
  Transformer on Time Series Forecasting.
\newblock In \emph{Advances in Neural Information Processing Systems},
  volume~32.

\bibitem[{Li et~al.(2020)Li, AbuFarha, Liu, Cheng, and Gall}]{li2020ms}
Li, S.-J.; AbuFarha, Y.; Liu, Y.; Cheng, M.-M.; and Gall, J. 2020.
\newblock MS-TCN++: Multi-Stage Temporal Convolutional Network for Action
  Segmentation.
\newblock \emph{IEEE Transactions on Pattern Analysis and Machine
  Intelligence}.

\bibitem[{Paszke et~al.(2019)Paszke, Gross, Massa, Lerer, Bradbury, Chanan,
  Killeen, Lin, Gimelshein, Antiga et~al.}]{pytorch}
Paszke, A.; Gross, S.; Massa, F.; Lerer, A.; Bradbury, J.; Chanan, G.; Killeen,
  T.; Lin, Z.; Gimelshein, N.; Antiga, L.; et~al. 2019.
\newblock Pytorch: An imperative style, high-performance deep learning library.
\newblock In \emph{Advances in Neural Information Processing Systems},
  8026--8037.

\bibitem[{Richard and Gall(2016)}]{richard2016temporal}
Richard, A.; and Gall, J. 2016.
\newblock Temporal action detection using a statistical language model.
\newblock In \emph{Proceedings of the IEEE Conference on Computer Vision and
  Pattern Recognition}, 3131--3140.

\bibitem[{Richard, Kuehne, and Gall(2017)}]{richard2017weakly}
Richard, A.; Kuehne, H.; and Gall, J. 2017.
\newblock Weakly supervised action learning with rnn based fine-to-coarse
  modeling.
\newblock In \emph{Proceedings of the IEEE Conference on Computer Vision and
  Pattern Recognition}, 754--763.

\bibitem[{Rohrbach et~al.(2012)Rohrbach, Amin, Andriluka, and
  Schiele}]{rohrbach2012database}
Rohrbach, M.; Amin, S.; Andriluka, M.; and Schiele, B. 2012.
\newblock A database for fine grained activity detection of cooking activities.
\newblock In \emph{2012 IEEE conference on computer vision and pattern
  recognition}, 1194--1201. IEEE.

\bibitem[{Ronneberger, Fischer, and Brox(2015)}]{ron2015unet}
Ronneberger, O.; Fischer, P.; and Brox, T. 2015.
\newblock U-net: Convolutional networks for biomedical image segmentation.
\newblock In \emph{International Conference on Medical image computing and
  computer-assisted intervention}, 234--241. Springer.

\bibitem[{Shaw, Uszkoreit, and Vaswani(2018)}]{shaw2018self}
Shaw, P.; Uszkoreit, J.; and Vaswani, A. 2018.
\newblock Self-attention with relative position representations.
\newblock \emph{arXiv preprint arXiv:1803.02155}.

\bibitem[{Singh et~al.(2016)Singh, Marks, Jones, Tuzel, and
  Shao}]{singh2016multi}
Singh, B.; Marks, T.~K.; Jones, M.; Tuzel, O.; and Shao, M. 2016.
\newblock A multi-stream bi-directional recurrent neural network for
  fine-grained action detection.
\newblock In \emph{Proceedings of the IEEE conference on computer vision and
  pattern recognition}, 1961--1970.

\bibitem[{Singhania, Rahaman, and Yao(2021)}]{singhania2021coarse}
Singhania, D.; Rahaman, R.; and Yao, A. 2021.
\newblock Coarse to fine multi-resolution temporal convolutional network.
\newblock \emph{arXiv preprint arXiv:2105.10859}.

\bibitem[{Stein and McKenna(2013)}]{stein2013combining}
Stein, S.; and McKenna, S.~J. 2013.
\newblock Combining embedded accelerometers with computer vision for
  recognizing food preparation activities.
\newblock In \emph{Proceedings of the 2013 ACM international joint conference
  on Pervasive and ubiquitous computing}, 729--738.

\bibitem[{Ulyanov, Vedaldi, and Lempitsky(2016)}]{ulyanov2016instance}
Ulyanov, D.; Vedaldi, A.; and Lempitsky, V. 2016.
\newblock Instance normalization: The missing ingredient for fast stylization.
\newblock \emph{arXiv preprint arXiv:1607.08022}.

\bibitem[{Van Den~Oord et~al.(2016)Van Den~Oord, Dieleman, Zen, Simonyan,
  Vinyals, Graves, Kalchbrenner, Senior, and Kavukcuoglu}]{van2016wavenet}
Van Den~Oord, A.; Dieleman, S.; Zen, H.; Simonyan, K.; Vinyals, O.; Graves, A.;
  Kalchbrenner, N.; Senior, A.~W.; and Kavukcuoglu, K. 2016.
\newblock WaveNet: A generative model for raw audio.
\newblock \emph{SSW}.

\bibitem[{Vaswani et~al.(2017)Vaswani, Shazeer, Parmar, Uszkoreit, Jones,
  Gomez, Kaiser, and Polosukhin}]{Transformer}
Vaswani, A.; Shazeer, N.; Parmar, N.; Uszkoreit, J.; Jones, L.; Gomez, A.~N.;
  Kaiser, {\L}.; and Polosukhin, I. 2017.
\newblock Attention is all you need.
\newblock In \emph{Advances in Neural Information Processing Systems}.

\bibitem[{Vo and Bobick(2014)}]{vo2014stochastic}
Vo, N.~N.; and Bobick, A.~F. 2014.
\newblock From stochastic grammar to bayes network: Probabilistic parsing of
  complex activity.
\newblock In \emph{Proceedings of the IEEE conference on computer vision and
  pattern recognition}, 2641--2648.

\bibitem[{Wang et~al.(2021)Wang, Xu, Wang, Shen, Cheng, Shen, and
  Xia}]{wang2021end}
Wang, Y.; Xu, Z.; Wang, X.; Shen, C.; Cheng, B.; Shen, H.; and Xia, H. 2021.
\newblock End-to-end video instance segmentation with transformers.
\newblock In \emph{Proceedings of the IEEE Conference on Computer Vision and
  Pattern Recognition}, 8741--8750.

\bibitem[{Wang et~al.(2020)Wang, Gao, Wang, Li, and Wu}]{BCN}
Wang, Z.; Gao, Z.; Wang, L.; Li, Z.; and Wu, G. 2020.
\newblock Boundary-Aware Cascade Networks for Temporal Action Segmentation.
\newblock In \emph{European Conference on Computer Vision}.

\bibitem[{Xie et~al.(2021)Xie, Wang, Yu, Anandkumar, Alvarez, and
  Luo}]{xie2021segformer}
Xie, E.; Wang, W.; Yu, Z.; Anandkumar, A.; Alvarez, J.~M.; and Luo, P. 2021.
\newblock SegFormer: Simple and efficient design for semantic segmentation with
  transformers.
\newblock \emph{Advances in Neural Information Processing Systems}, 34.

\bibitem[{Xu et~al.(2021)Xu, Wu, Wang, and Long}]{xu2021anomaly}
Xu, J.; Wu, H.; Wang, J.; and Long, M. 2021.
\newblock Anomaly Transformer: Time Series Anomaly Detection with Association
  Discrepancy.
\newblock \emph{arXiv preprint arXiv:2110.02642}.

\bibitem[{Yan et~al.(2021)Yan, Peng, Fu, Wang, and Lu}]{yan2021learning}
Yan, B.; Peng, H.; Fu, J.; Wang, D.; and Lu, H. 2021.
\newblock Learning spatio-temporal transformer for visual tracking.
\newblock \emph{arXiv preprint arXiv:2103.17154}.

\bibitem[{Yi, Wen, and Jiang(2021)}]{ASFormer}
Yi, F.; Wen, H.; and Jiang, T. 2021.
\newblock ASFormer: Transformer for Action Segmentation.
\newblock In \emph{The British Machine Vision Conference}.

\bibitem[{Zhang et~al.(2019)Zhang, Muandet, Ma, Neumann, and
  Tang}]{zhang2019low}
Zhang, Y.; Muandet, K.; Ma, Q.; Neumann, H.; and Tang, S. 2019.
\newblock Low-rank random tensor for bilinear pooling.
\newblock \emph{arXiv}, arXiv--1906.

\bibitem[{Zhou et~al.(2021)Zhou, Zhang, Peng, Zhang, Li, Xiong, and
  Zhang}]{informer}
Zhou, H.; Zhang, S.; Peng, J.; Zhang, S.; Li, J.; Xiong, H.; and Zhang, W.
  2021.
\newblock Informer: Beyond Efficient Transformer for Long Sequence Time-Series
  Forecasting.
\newblock In \emph{The Thirty-Fifth AAAI Conference on Artificial
  Intelligence}.

\end{thebibliography}


\clearpage
\appendix

\section{Appendix A More Ablations}
In this section, we replace all the local attention modules in the TUT model with 1D convolutions to illustrate the superiority of attention over convolutions (see Table~\ref{tab:conv}). We also conduct more ablation studies about more hyper-parameters in the multi-head local attention: the window size $w$ (see Table~\ref{tab:w}) and the number of heads $h$ (see Table~\ref{tab:h}). Since the boundary-aware loss is obtained by measuring the distance between the prior distribution and the local-attention distribution, the effect of which is controlled by the weight $\beta$, we also study other distribution metrics (see Table~\ref{tab:baloss}) and the hyper-parameter $\beta$ (see Table~\ref{tab:beta}). All the ablation studies are performed on the
50Salads dataset.

\subsection{Attention vs 1D Convolutions}
We replace all the local attention modules of window size $w$ in the TUT model with 1D convolutions of kernel size $w$. For each encoder layer, the number of input and output channels of the convolutions are both $d$, where $d$ is the hidden dimension. For each decoder layer, following U-Net, we pass the information from the encoder to the decoder by concatenating the encoder layer output instead of cross attention. Therefore, the number of input channels of the convolutions is $2d$, while the number of output channels is $d$. We find that the performance of TUT with local attention is much better than that with convolutions, which illustrates the limitations of 1D convolutions and the superiority of attention.

\begin{table}[ht]
\centering
\begin{tabular}{cccccc}
    \toprule
    Module   & \multicolumn{3}{c}{F1@\{10,25,50\}} & Edit & Acc\\
    \midrule
    Attention & \textbf{87.7} & \textbf{87.1} & \textbf{79.9} & \textbf{82.6} & \textbf{85.9}\\
    Convolution & 82.8 & 81.3 & 74.1 & 75.5 & 85.2\\
    \bottomrule
\end{tabular}
\caption{Comparison of attention and 1D convolutions.}
\label{tab:conv}
\end{table}

\subsection{Impact of \(w\) and \(h\)}
Since the parameter amount of the embedding matrix $W_{rpe}\in\mathbb{R}^{w\times h}$ in RPE depends on $w$ and $h$, we do not use RPE in these ablations for a fair comparison. If $w$ is small, the receptive field is too small to capture long-term dependencies between frames; and if $w$ is too large, the model is difficult to be trained adequately on small-scale data. Therefore, it is best to set $w$ to a number that is neither too large nor too small. In Table~\ref{tab:w}, the model performs best when $w=51$. 

\begin{table}[ht]
\centering
\begin{tabular}{cccccc}
    \toprule
    $w$    & \multicolumn{3}{c}{F1@\{10,25,50\}} & Edit & Acc\\
    \midrule
    11 & 78.9 & 77.0 & 70.1 & 71.0 & 84.0\\
    31 & 84.3 & 83.0 & 75.6 & 77.3 & 85.8\\
    51 & 86.4 & 85.4 & 78.6 & 80.5 & 85.4\\
    71 & 85.7 & 84.2 & 78.2 & 79.5 & 84.7\\
    91 & 85.2 & 83.5 & 76.7 & 79.3 & 84.2\\
    \bottomrule
\end{tabular}
\caption{Effect of the window size $w$ in local attention.}
\label{tab:w}
\end{table}

We observe that the model performs better as the number of heads $h$ increases in Table~\ref{tab:h}. Since each head maintains an attention matrix, the memory consumption also increases with the number of heads. Considering that the performance gain brought by the additional memory consumption is not obvious, we set $h=4$ by default on the 50Salads dataset.

\begin{table}[ht]
\centering
\begin{tabular}{ccccccc}
    \toprule
    $h$    & \multicolumn{3}{c}{F1@\{10,25,50\}} & Edit & Acc & GPU Mem.\\
    \midrule
    1 & 85.1 & 83.2 & 74.3 & 78.0 & 83.3 & 3.0G\\
    2 & 84.7 & 83.5 & 75.6 & 78.1 & 84.3 & 3.1G\\
    4 & 86.4 & 85.4 & 78.6 & 80.5 & 85.4 & 3.5G\\
    8 & 87.0 & 86.1 & 79.2 & 80.9 & 85.7 & 5.0G\\
    \bottomrule
\end{tabular}
\caption{Effect of the number of heads $h$ in multi-head local attention.}
\label{tab:h}
\end{table}

\subsection{Impact of \(\beta\)}
According to Table~\ref{tab:beta}, we set $\beta=0.02$ on the 50Salads dataset. When reducing the value of $\beta$ to 0.01, the performance improvement is not obvious. Larger values of $\beta$ cause performance degradation, the reason of which is that the transitional focus on boundary frames reduces the classification accuracy of intermediate frames.

\begin{table}[ht]
\centering
\begin{tabular}{cccccc}
    \toprule
    $\beta$    & \multicolumn{3}{c}{F1@\{10,25,50\}} & Edit & Acc\\
    \midrule
    0 & 87.7 & 87.1 & 79.9 & 82.6 & 85.9\\
    0.01 & 88.2 & 87.5 & 80.6 & 83.1 & 86.2\\
    0.02 & \textbf{89.3} & \textbf{88.3} & \textbf{81.7} & \textbf{84.0} & \textbf{87.2}\\
    0.03 & 88.1 & 87.8 & 80.6 & 83.5 & 86.6\\
    0.04 & 87.2 & 86.5 & 79.3 & 82.1 & 85.0\\
    \bottomrule
\end{tabular}
\caption{Effect of $\beta$ which is the weight of the proposed boundary-aware loss.}
\label{tab:beta}
\end{table}

\subsection{Impact of distance metrics}
We select the following widely-used statistical distances to calculate the boundary-aware loss:
\begin{itemize}
\item Kullback–Leibler Divergence (KL, Ours).
\item Jensen–Shannon Divergence (JS).
\item Wasserstein Distance (Wasserstein).
\item L2 Distance (L2).
\end{itemize}

As shown in Table~\ref{tab:baloss}, an inappropriate distance metric can even hurt performance. The prior distribution is predefined and fixed, and we expect the local-attention distribution to be similar to it, so we do not need a symmetric Jensen–Shannon divergence. Since the distribution areas are completely overlapping, i.e., within a local window, the Wasserstein distance does not play to its advantage. Instead, the Kullback–Leibler divergence we adopt performs best.

\begin{table}[ht]
\centering
\begin{tabular}{cccccc}
\toprule
Metric & \multicolumn{3}{c}{F1@\{10,25,50\}} & Edit  & Acc \\
\midrule
\ding{56}  & 87.7 & 87.1 & 79.9 & 82.6 & 85.9\\
\midrule
L2  & 88.6 & 87.4 & 80.8 & 82.6 & 86.9\\
Wasserstein & 87.5 & 86.2 & 79.2 & 81.9 & 85.6\\
JS & 87.6 & 86.4 & 80.0 & 81.7 & 86.4 \\
\midrule 
KL (Ours) & \textbf{89.3} & \textbf{88.3} & \textbf{81.7} & \textbf{84.0} & \textbf{87.2}\\
\bottomrule
\end{tabular}
\caption{Model performance under different distribution distance metrics in the boundary-aware loss.}
\label{tab:baloss}
\end{table}

\section{Appendix B Qualitative Results}
\subsection{Comparison with ASFormer}
\label{sec:results}
\begin{figure}[ht]
\begin{center}
\subfigure{\includegraphics[width=1\columnwidth]{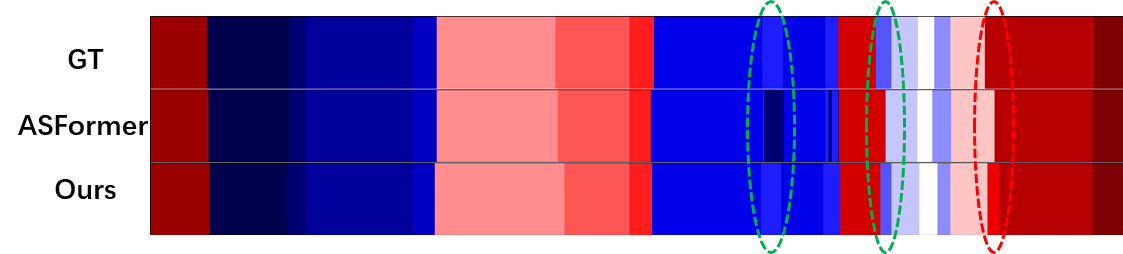}}
\subfigure{\includegraphics[width=1\columnwidth]{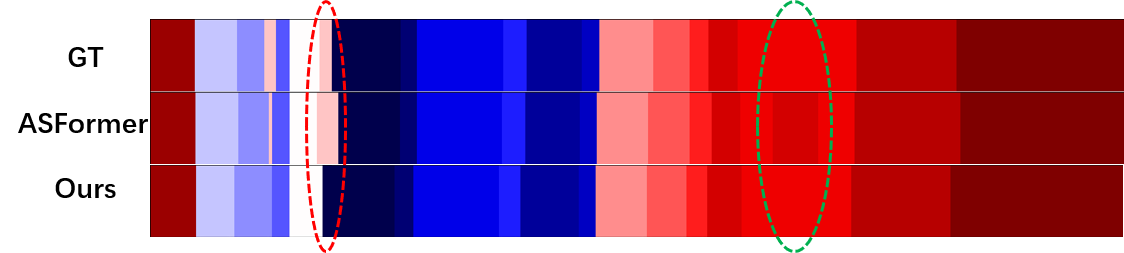}}   
\subfigure{\includegraphics[width=1\columnwidth]{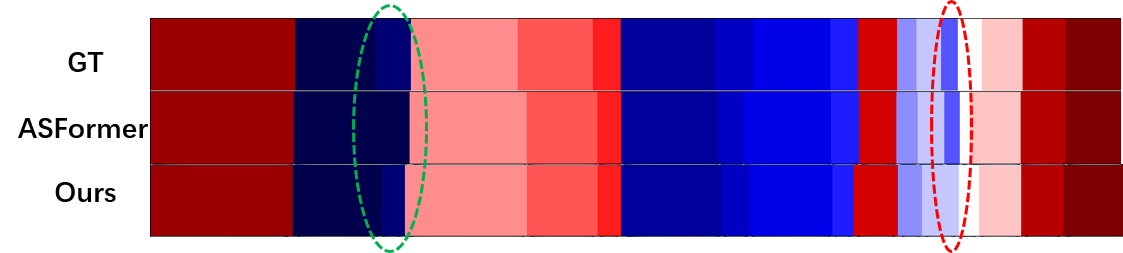}} 
\subfigure{\includegraphics[width=1\columnwidth]{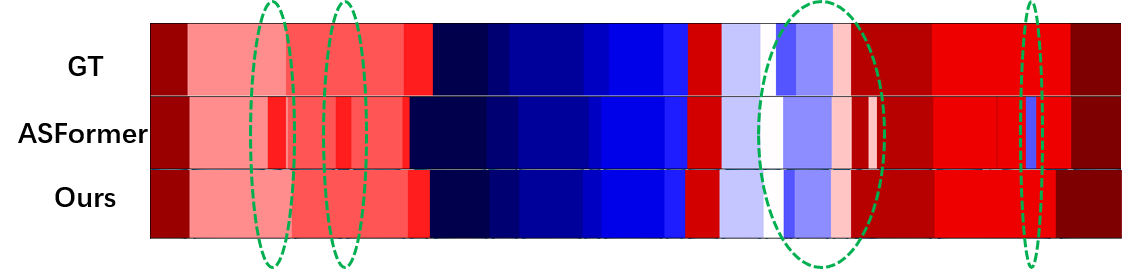}}
\caption{Visualization of the segmentation results of some videos, where the GT row represents the ground truth, the ASFormer row represents the predictions of the ASFormer, and Ours represents the predictions of our model. The same color represents the same action class.}
\label{fig:sota}
\end{center}
\end{figure}

ASFormer is the best backbone model before, so we compare its predictions and ours in Fig.~\ref{fig:sota}. Our purpose is to fairly compare the performance of backbone models, so we show results of our model which is trained without the boundary-aware loss. The action segments in the green circle are examples that our model classifies correctly but ASFormer misclassifies, while the action segments in the red circle are the opposite. Overall, our model works much better. For those very short action segments (compared to the entire video length), our model may ignore them. And for other cases, our model performs reasonably well.

\subsection{The Effect of Boundary-Aware Loss}
\begin{figure}[ht]
\begin{center}
\subfigure{\includegraphics[width=1\columnwidth]{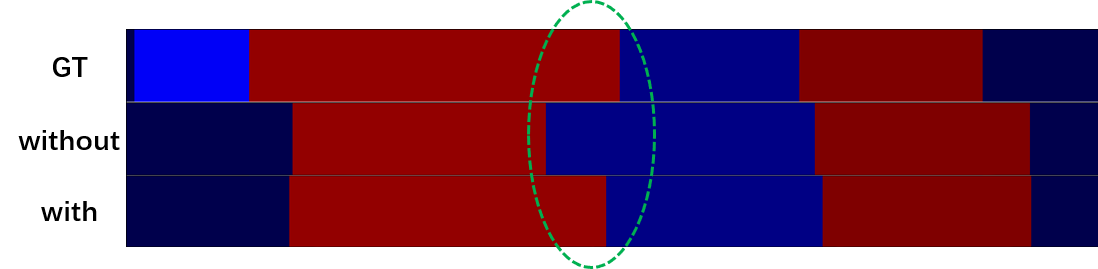}}
\subfigure{\includegraphics[width=1\columnwidth]{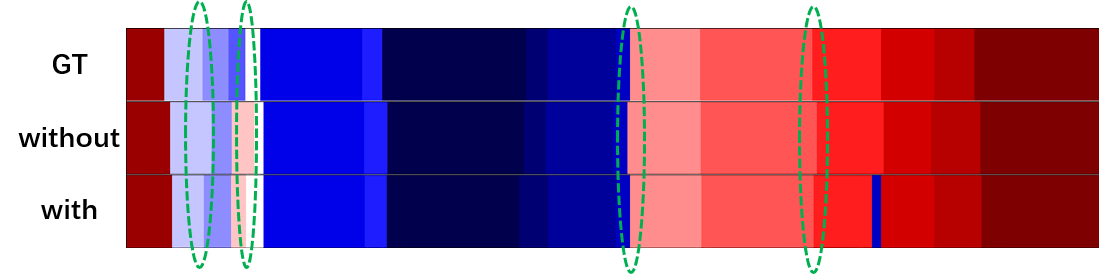}}   
\subfigure{\includegraphics[width=1\columnwidth]{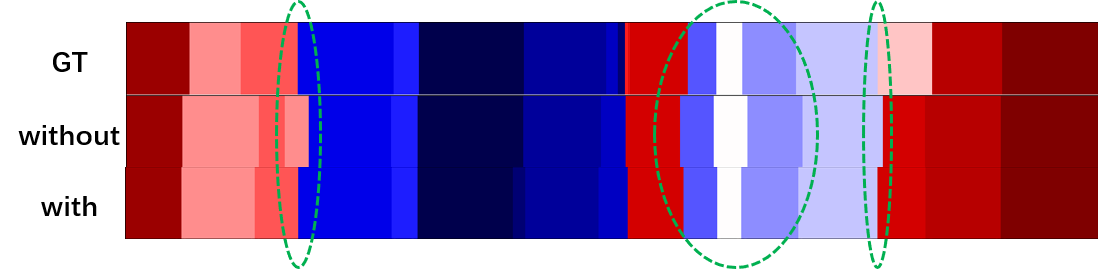}} 
\subfigure{\includegraphics[width=1\columnwidth]{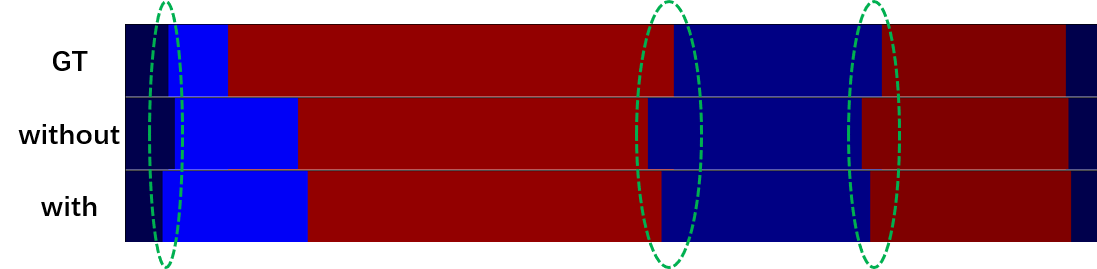}}
\caption{Visualization of the segmentation results of some videos, where the GT row represents the ground truth, the \textit{with} row and \textit{without} row represent the predictions of our EUT which is trained with and without the boundary-aware loss, respectively. The same color represents the same action class.}
\label{fig:balosseffect}
\end{center}
\end{figure}

We show more visual comparisons of our model which is trained with or without the boundary-aware loss in Fig.~\ref{fig:balosseffect}. Comparing the results in the green circle, we observe that the boundary frames predicted by the model trained with boundary-aware loss are closer to the real boundary frames.

\section{Appendix C Experiment Details}

\begin{table*}[t]
\centering
\begin{tabular}{cccccc}
    \toprule
    Datasets & \#Sample & Length & \#Class & \#Split & Scene \\
    \midrule
    50Salads & 50 & 11552 & 17 & 5 & preparing salads \\
    GTEA  & 28 & 1115 & 11 & 4 & daily activities \\
    Breakfast & 1712 & 2097 & 48 & 4& cooking breakfast \\
    \bottomrule
\end{tabular}
\caption{Details of three datasets. \#Sample and \#Class are the numbers of video samples and classes, respectively. \#Split is the numer of splits. Length is the average frames of videos.}
\label{tab:datasets}
\end{table*}

\begin{table*}[t]
\centering
\begin{tabular}{cccc}
    \toprule
    & 50Salads & GTEA  & Breakfast \\
    \midrule
    \textbf{\#Refinement Stages} $M$                & 3 & 3 & 3 \\
    \textbf{\#Layers} $N$                & 5 &  4 & 5 \\
    \textbf{Window Size} $w$                & 51 & 11 & 25 \\
    \textbf{Hidden Dimension in the Prediction stage}   & 128 & 64 &  192\\
    \textbf{FFN Inner Dimension in the Prediction stage}     & 128 & 64 &  192\\
    \textbf{Hidden Dimension in Refinement stages}   & 64 &  64 & 96\\
    \textbf{FFN Inner Dimension in Refinement stages} & 64 &  64 & 96\\
    \textbf{\#Attention Heads} $h$           & 4 &  4 & 6\\
    \textbf{Input Dropout}  & 0.4 & 0.5  & 0.4\\
    \textbf{FFN Dropout}    & 0.3 & 0.3 &  0.3\\
    \textbf{Attention Dropout}    & 0.2 &  0.2 &  0.2\\
    \textbf{Learning Rate}   & 5e-4 & 5e-4 &  2e-4\\
    \textbf{Weight Decay}   & 1e-5 & 1e-5 &  5e-5\\
    \textbf{Boundary-Aware Loss Weight $\beta$}   & 0.02 & 0.1 &  0.005\\
    \bottomrule
\end{tabular}
\caption{Model configurations of EUT and training hyper-parameters on three datasets.}
\label{tab:hyperparameters}
\end{table*}

\subsection{Details of Datasets}
\label{sec:dataset}
We list the details of three action segmentation datasets in Table~\ref{tab:datasets}. The GTEA dataset has the smallest number of samples and the shortest average length. The 50Salads dataset has the longest average video length. And the Breakfast dataset has the largest number of samples. Therefore, the latter two datasets will provide more supervised signals during training and are more conducive to training our pure Transformer model.

\subsection{Details of Settings}
For all experiments, we employ a learning rate decay strategy, which reduces the learning rate by half when the training loss increases three times. In Table~\ref{tab:hyperparameters}, we report the model configurations and training hyper-parameters corresponding to the experimental results in the paper. Following MS-TCN and ASFormer, we select the epoch number that can achieve the best average result for all splits to report the results on every dataset. On 50Salads, the sampling rate is reduced from 30 fps to 15 fps, and the output is up-sampled to 30 fps. We vary the random seed to repeat each experiment three times and report the mean of the metrics in the paper.

\end{document}